\documentclass[letterpaper, 10 pt]{IEEEtran}
\usepackage{graphics} 
\usepackage{graphicx} 
\usepackage{epsfig} 
\usepackage{amsmath, bm} 
\usepackage{amssymb}  
\usepackage{upgreek}
\usepackage{multicol}
\usepackage{multirow}
\usepackage{lipsum}  
\usepackage{breqn}
\usepackage{comment} 
\usepackage{booktabs}

\PassOptionsToPackage{hyphens}{url}\usepackage{hyperref}
\usepackage[caption=false]{subfig} 

\ifCLASSINFOpdf
\else
\fi
\begin{document}
%
\title{Mechanism Design of a Bio-inspired Armwing Mechanism for Mimicking Bat Flapping Gait}
%
%
%

\author{Eric Sihite$^{1}$, Peter Kelly$^{1}$, and Alireza Ramezani$^{1}$%
\thanks{$^{1}$The authors are with the SiliconSynapse Laboratory, Department of Electrical and Computer Engineering, Northeastern University, Boston, MA 02119, USA (e-mail: ericsihite@gmail.com; kelly.pe@husky.neu.edu, a.ramezani@northeastern.edu).}%
}
\maketitle

\IEEEpeerreviewmaketitle


\section{Abstract}
\label{sec:introduction}

\begin{figure*}[t]
    \vspace{0.05in}
    \centering
    \includegraphics[width=0.9\linewidth]{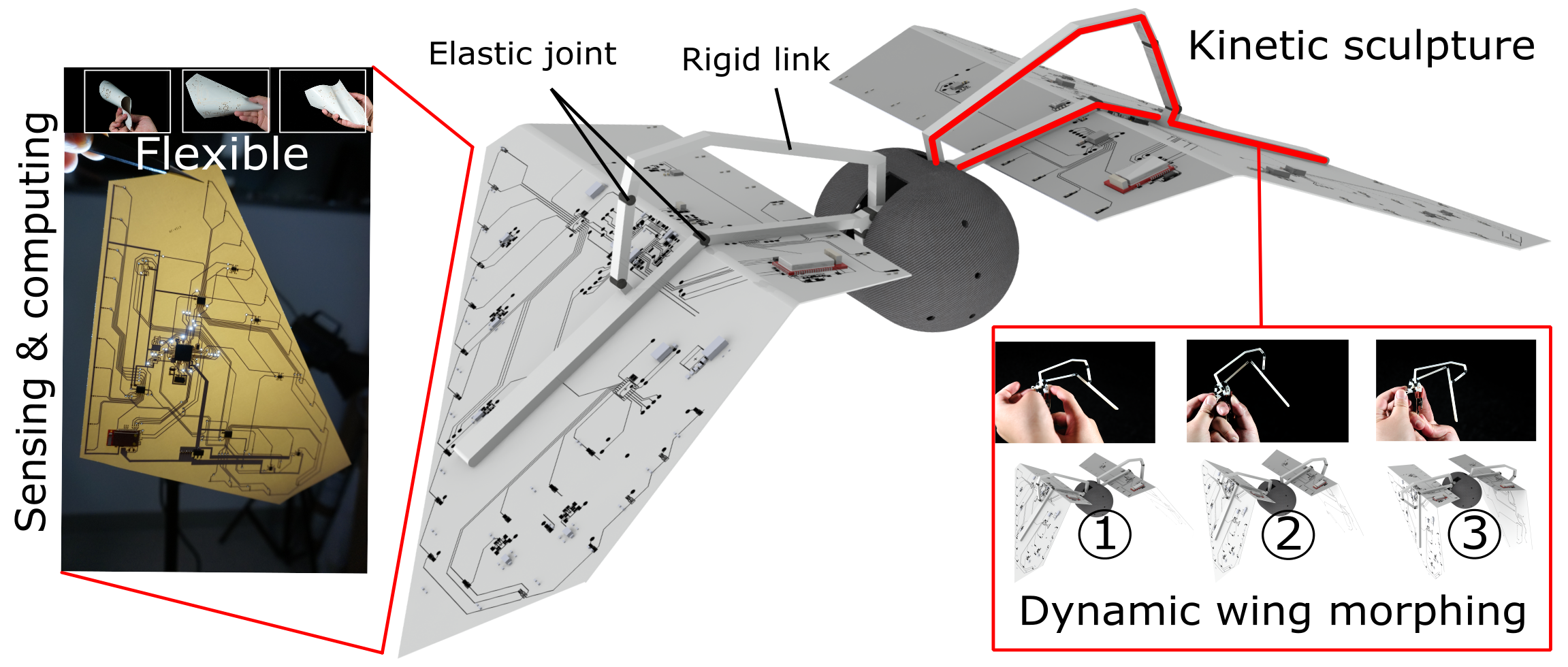}
    \caption{The bio-inspired bat robot, called \textit{Aerobat}, uses monolithically fabricated rigid and soft arm-armwing structure. The white material of the armwing is rigid while the black material is a flexible living hinge. The flexible PCB forms the wing membrane which helps reducing the overall weight of the robot and the \textit{kinetic sculpture} forms a compliant linkage mechanism that can conform and articulate flapping motion.}
    \label{fig:satan_overview}
    \vspace{-0.05in}
\end{figure*}


The overall goal of this work is to advance the theory and practice of aerial robots that are soft, agile, collision-tolerant, and energetically efficient by the biomimicry of key airborne vertebrate flight characteristics. In recent years, much attention has been drawn to make our residential and work spaces smarter and to materialize the concept of smart cities \cite{everaerts2008use}. As a result, safety and security aspects are gaining ever growing importance \cite{pavlidis2001urban} and drive a lucrative market. Systems that can provide situational awareness to humans in residential and work spaces or contribute to dynamic traffic control in cities will result in large-scale intelligent systems with enormous societal impact and economic benefit. 

Current state-of-the-art solutions with rotary or fixed-wing features fall short in addressing the challenges and pose extreme dangers to humans. Fixed or rotary-wing systems are widely applied for surveillance and reconnaissance, and there is a growing interest to add suites of on-board sensors to these unmanned aerial systems (UAS) and use their aerial mobility to monitor and detect hazardous situations in residential spaces. While, these systems, e.g., quadrotors, can demonstrate agile maneuvers and have demonstrated impressive fault-tolerance in aggressive environments, quadrotors and other rotorcrafts require a safe and collision-free task space for operation since they are not collision-tolerant due to their rigid body structures. The incorporation of soft and flexible materials into the design of such systems has become common in recent years, yet, the demands for aerodynamic efficiency prohibit the use of rotor blades made of flexible materials.

The flight apparatus of birds and bats can offer invaluable insights into novel micro aerial vehicle (MAV) designs that can safely operate within residential spaces. The pronounced body articulation (morphing ability) of these flyers is key to their unparalleled capabilities. These animals can reduce the wing area during upstrokes and can extend it during downstrokes to maximize positive lift generation \cite{tobalske2000biomechanics}. It is known that some species of bats can use differential inertial forces to perform agile zero-angular momentum turns \cite{riskin2012upstroke}. Biological studies suggest that the articulated musculoskeletal system of animals can absorb impact forces therefore can enhance their survivability in the event of a collision \cite{roberts2013tendons}.

The objective of this work is to design and develop a bio-inspired soft and articulated armwing structure which will be an integral component of a morphing aerial co-bot, \textit{Aerobat}, which is shown in Fig. \ref{fig:satan_overview}. In our design, we draw inspiration from bats. Bat membranous wings possess unique functions \cite{tanaka2015flexible} that make them a good example to take inspiration from and transform current aerial drones. In contrast with other flying vertebrates, bats have an extremely articulated musculoskeletal system, key to their body impact-survivability and deliver an impressively adaptive and multimodal locomotion behavior \cite{riskin2008quantifying}. Bats exclusively use this capability with structural flexibility to generate the controlled force distribution on each wing membrane. The wing flexibility, complex wing kinematics, and fast muscle actuation allow these creatures to change the body configuration within a few tens of milliseconds. These characteristics are crucial to the unrivaled agility of bats \cite{azuma2006biokinetics} and copying them can potentially transform the state-of-the-art aerial drone design. 

\begin{figure}[t]
    \centering
    \includegraphics[width=0.6\linewidth]{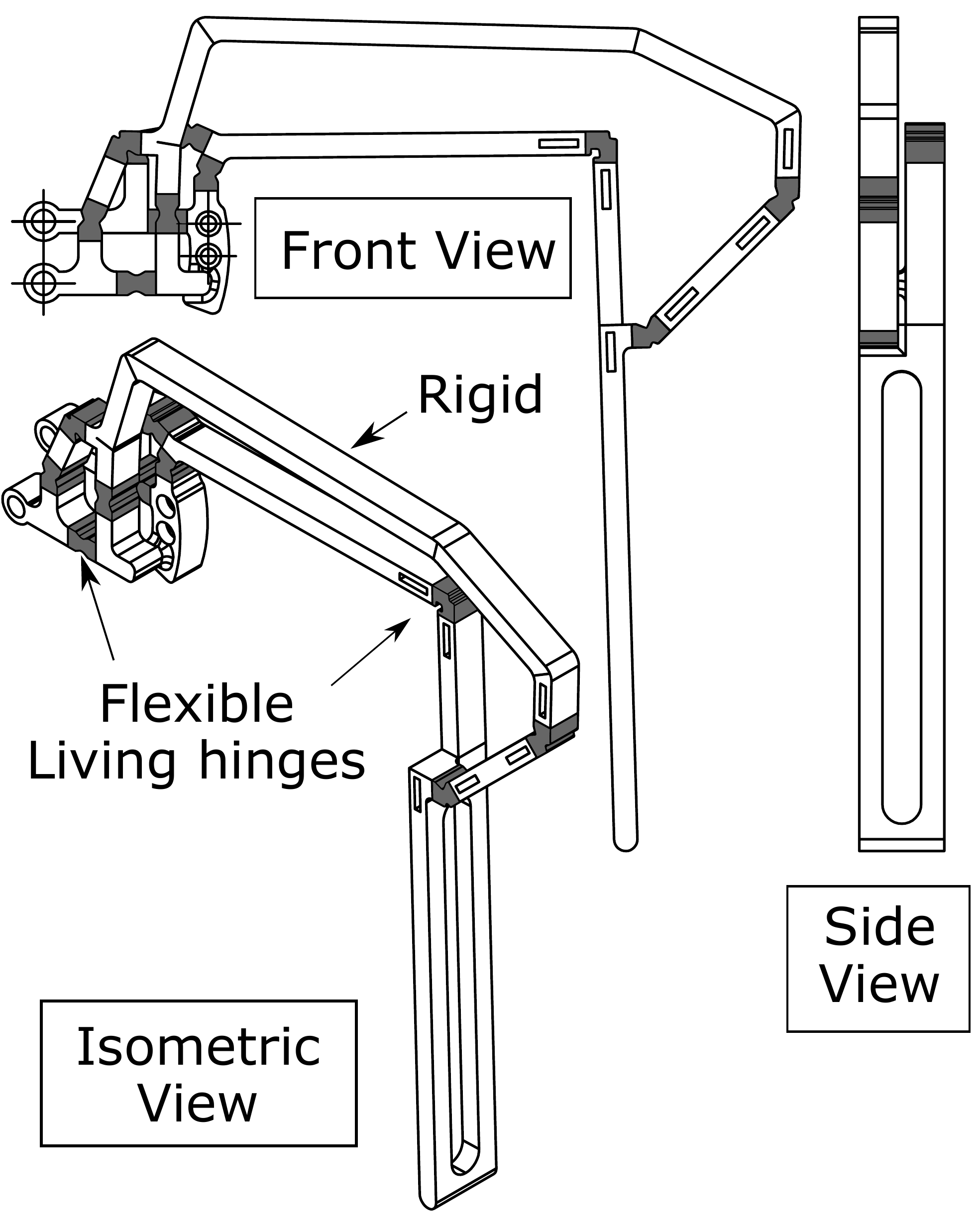}
    \caption{The CAD model of the monolithic wing structure for PolyJet 3D printing. The white and greyed sections are rigid and flexible hinges respectively which are made using a mixture between the rigid \textit{Vero White} and the flexible \textit{Agilus Black} (Stratasys polyjet materials).}
    \label{fig:wing_cad}
\end{figure}


The emerging ideas surrounding achieving computation in robots through sophisticated interactions of morphology, called \textit{morphological computation} or \textit{mechanical intelligence} \cite{hauser2012role}, draws our attention to the obvious fact that there is a common interconnection -- and in some morphologies these couplings are very tight -- between the boundaries of morphology and closed-loop feedback. Controllers lie in the space of abstract computation, and are usually implemented in computational layers or are programmed into the system. However, if mechanical interactions can also perform computation, it becomes possible for the morphology to play a role of computation in the system, and in effect part of the role of the controller is subsumed under computational morphology. Particularly, we will explore such design approaches to copy dynamically versatile wing conformations of bats flight apparatus. 

This work extends our prior contributions \cite{ramezani2016bat,hoff2016synergistic,hoff2018optimizing,hoff2017reducing,ramezani2017biomimetic} where we attempted to design armwing retraction mechanisms and used the opportunity to study the underlying control mechanisms \cite{ramezani_describing_2017,ramezani_lagrangian_2015,ramezani_modeling_2016,hoff_trajectory_2019,syed_rousettus_2017} based on which bats perform sharp banking turns and diving maneuvers. We improve our previous design by developing kinetic sculpture designs that can capture bat dynamically versatile wing conformations. These structures consist of rigid and flexible materials that are monolithically fabricated using novel computed-aided fabrication methods and additive manufacturing technology (PolyJet 3D printing), as shown in Fig. \ref{fig:wing_cad}. Like its predecessor, this armwing structure articulation is also designed to expand and retract within a single wingbeat through a series of crank and four-bar mechanisms as it is actuated by a single brushless DC motor. The use of a monolithic rigid and flexible armwing structure in a flying robot is novel and might be very impactful for flapping robot design as this structure is capable of mimicking the range of motion and flexibility of an actual bat armwing. This mechanism design assumes a planar flapping motion and articulates the wing plunging and extension-retraction gaits.

\ifCLASSOPTIONcaptionsoff
  \newpage
\fi




%

\bibliographystyle{IEEEtran}
\bibliography{references_local}




%








\end{document}